\documentclass[10pt,twocolumn,letterpaper]{article}

\usepackage{cvpr}              

\usepackage[dvipsnames]{xcolor}

\definecolor{cvprblue}{rgb}{0.21,0.49,0.74}
\usepackage[pagebackref,breaklinks,colorlinks,citecolor=cvprblue]{hyperref}
\usepackage{multirow, multicol}
\newcommand{\john}[1]{\textcolor{black}{#1}}

\def\paperID{10} 
\def\confName{CVPR}
\def\confYear{2024}

\title{ReduceFormer: Attention with Tensor Reduction by Summation}

\author{John Yang, Le An, Su Inn Park\\
NVIDIA\\
{\tt\small \{johnyang, lean, joshp\}@nvidia.com}
}

\begin{document}
\maketitle
\begin{abstract}

Transformers have excelled in many tasks including vision. However, efficient deployment of transformer models in low-latency or high-throughput applications is hindered by the computation in the attention mechanism which involves expensive operations such as matrix multiplication and Softmax.
To address this, we introduce \textit{ReduceFormer}, a family of models optimized for efficiency with the spirit of attention. ReduceFormer leverages only simple operations such as reduction and element-wise multiplication, leading to greatly simplified architecture and improved inference performance, with up to 37\% reduction in latency and 44\% improvement in throughput, while maintaining competitive accuracy comparable to other recent methods. The proposed model family is suitable for edge devices where compute resource and memory bandwidth are limited, as well as for cloud computing where high throughput is sought after. Code will be available soon.
\end{abstract}    
\section{Introduction}
\label{sec:intro}
Transformer networks, critical for advancing natural language processing and computer vision, utilize self-attention modules to efficiently process sequences and capture complex dependencies and global context directly \cite{liu2021swin, liu2022swin, waswani2017attention, dosovitskiy2020image_vit}. Unlike prior architectures based on recurrent or convolutional layers \cite{he2016deep_resnet, liu2022convnext, tan2019efficientnet}, transformers dynamically weigh the significance of different parts of the input data, facilitating powerful representation learning. However, the quadratic computational scaling of self-attention, alongside matrix multiplications and Softmax computation, requires substantial computational resources and memory, in turn, challenging efficient deployment such as on edge devices. This has spurred research into more efficient computational strategies to reduce reliance on these intensive processes \cite{cai2023efficientvit, yang2022depth, li2021efficient, jia2021efficient, wang2020linformer, wang2021pyramid, touvron2021training_deit}.

While self-attention mechanisms enable global interactions in transformers, they are not the only method. As an example, Wang \textit{et al.} \cite{wang2018non} introduce non-local operations, inspired by non-local means in image processing, which compute a position's response as a weighted sum of features across all positions. This approach effectively captures long-range dependencies, highlighting the potential of non-local methods for tasks like object detection and recognition in computer vision.

In this paper, we propose \textit{ReduceFormer}, a family of vision models by exclusively harnessing only basic operations such as element-wise multiplication and global summation to model both local and global feature relationships. The advantages of the proposed approach over previous methods are mainly twofold:

\begin{itemize}
\item \textbf{Reduced Model Complexity:} By our design choice, we eliminate the use of matrix multiplication and expensive operations such as Softmax as in typical attention blocks, leading to a much simpler model structure.

\item \textbf{Efficiency in Inference:} The operations in the proposed series of models can utilize well-optimized implementations on modern deep learning accelerators such as GPU, resulting in improved efficiency in latency, throughput, and memory footprint.


\end{itemize}
The proposed models exhibits significant speedup as compared to prior SOTA methods with competitive accuracy, on both embedded device and data center GPU.


\begin{figure*}[t]
\centering
\includegraphics[width=0.95\textwidth]{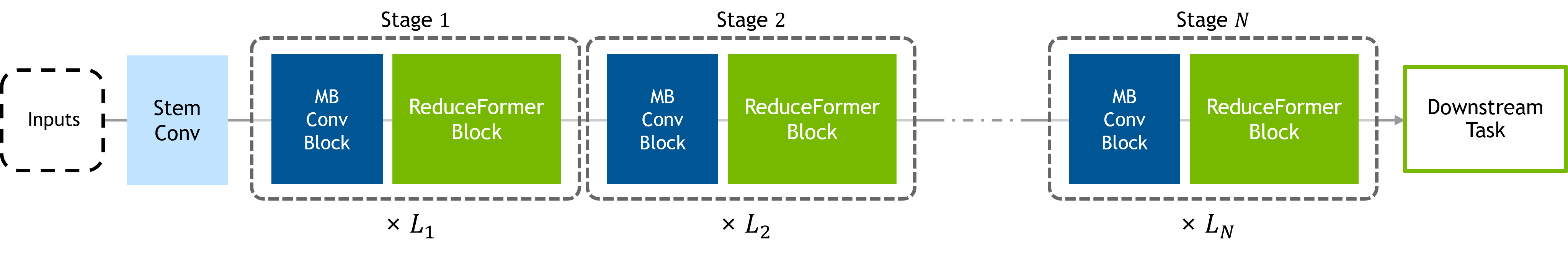}
\vspace{-3mm}
\caption{The proposed architecture of ReduceFormer. Each stage can be composed with $L$ repetitions of MB Conv blocks~\cite{sandler2018mobilenetv2} with/without ReduceFormer blocks.}
\vspace{-4mm}
\centering
\label{fig:whole_structure}
\end{figure*}


\section{Related Works}
\label{sec:related_works}
Built on top of the seminal work of attention-based transformer model~\cite{waswani2017attention}, the advent of Vision Transformer (ViT) has marked a significant milestone in the field of computer vision, with the pioneering work \cite{dosovitskiy2020image_vit} demonstrating that using a transformer model and processing image patches as sequence of tokens could rival the performance of Convolutional Neural Networks (CNNs) \cite{he2016deep_resnet, liu2022convnext}.

Efforts to enhance transformer efficiency have focused on reducing computational overhead, with theoretical and practical innovations \cite{choromanski2020rethinking, wang2020linformer}. For instance, Li \textit{et al.} introduced EsViT, a multi-stage Efficient Transformer utilizing sparse self-attention to decrease complexity \cite{li2021efficient}. Jia \textit{et al.} developed an efficient manifold distillation technique that allows a student network to outperform its teacher by refining transformer learning processes \cite{jia2021efficient}. However, these advancements often confront practical hardware limitations. Additionally, efforts to simplify transformer structures have led to replacing Softmax with max-pooling for enhanced efficiency in dense prediction tasks \cite{yang2022depth}, and EfficientViT modifies attention computations to streamline processes using a matrix product of queries \textit{Q} and keys \textit{K} followed by ReLU activation \cite{cai2023efficientvit}. Despite these improvements, challenges with expensive operations such as matrix multiplications during inference persist. Our method sidesteps these costly operations by focusing on simpler, more efficient ones, improving model performance without heavy computational demands.
\begin{figure}[!t]
\centering
\includegraphics[width=0.8\linewidth]{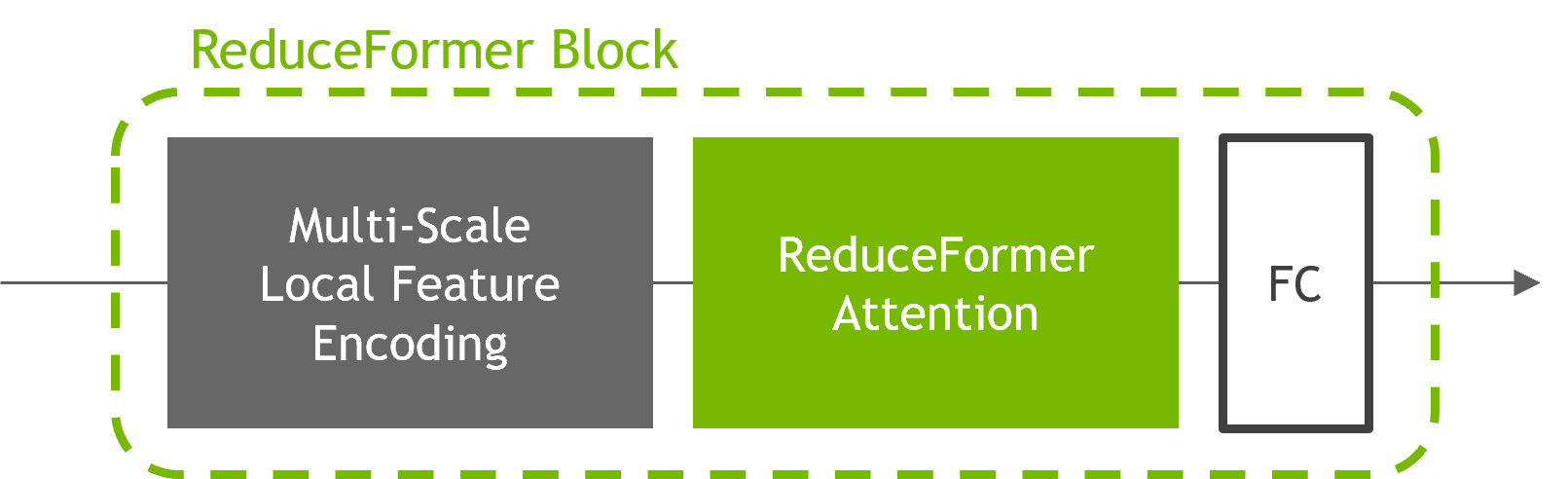}
\vspace{-2mm}
\caption{A ReduceFormer block consists of two phases: local context learning and global context learning via ReduceFormer attention.}
\vspace{-4.5mm}
\centering
\label{fig:RF_block}
\end{figure}

\section{Method}

The architecture of the proposed ReduceFormer is illustrated in Figure \ref{fig:whole_structure}. A block of \textit{Stem Conv} in the beginning of the network reduces the spatial dimensions of inputs and transitions their representation to a higher-dimensional channel space. Then, stacks of an inverted residual block \cite{sandler2018mobilenetv2} with/without a \textit{ReduceFormer} block are repeated $N$ times for hierarchical feature embedding from input. 

Our proposed method integrates an attention mechanism into the ReduceFormer block, designed to complement the capability of convolution operations to encapsulate global information. In pursuit of efficiency during model execution, we incorporate straightforward feature-reduction operations, such as global summation and element-wise operations, resulting in the designation of our model as \textit{ReduceFormer}. 
This approach is mainly characterized by ReLU-based 
attention processes, involving tensor reductions through global summations, followed by a Fully Connected (FC) layer, as illustrated in Figure \ref{fig:RF_block}.
This sequence prioritizes the extraction of local context information initially, achieved through the employment of a multi-scale local context learning block. Subsequently, the ReduceFormer Attention captures non-local feature relations.

\begin{figure}[!t]
\centering
\includegraphics[width=0.77\linewidth]{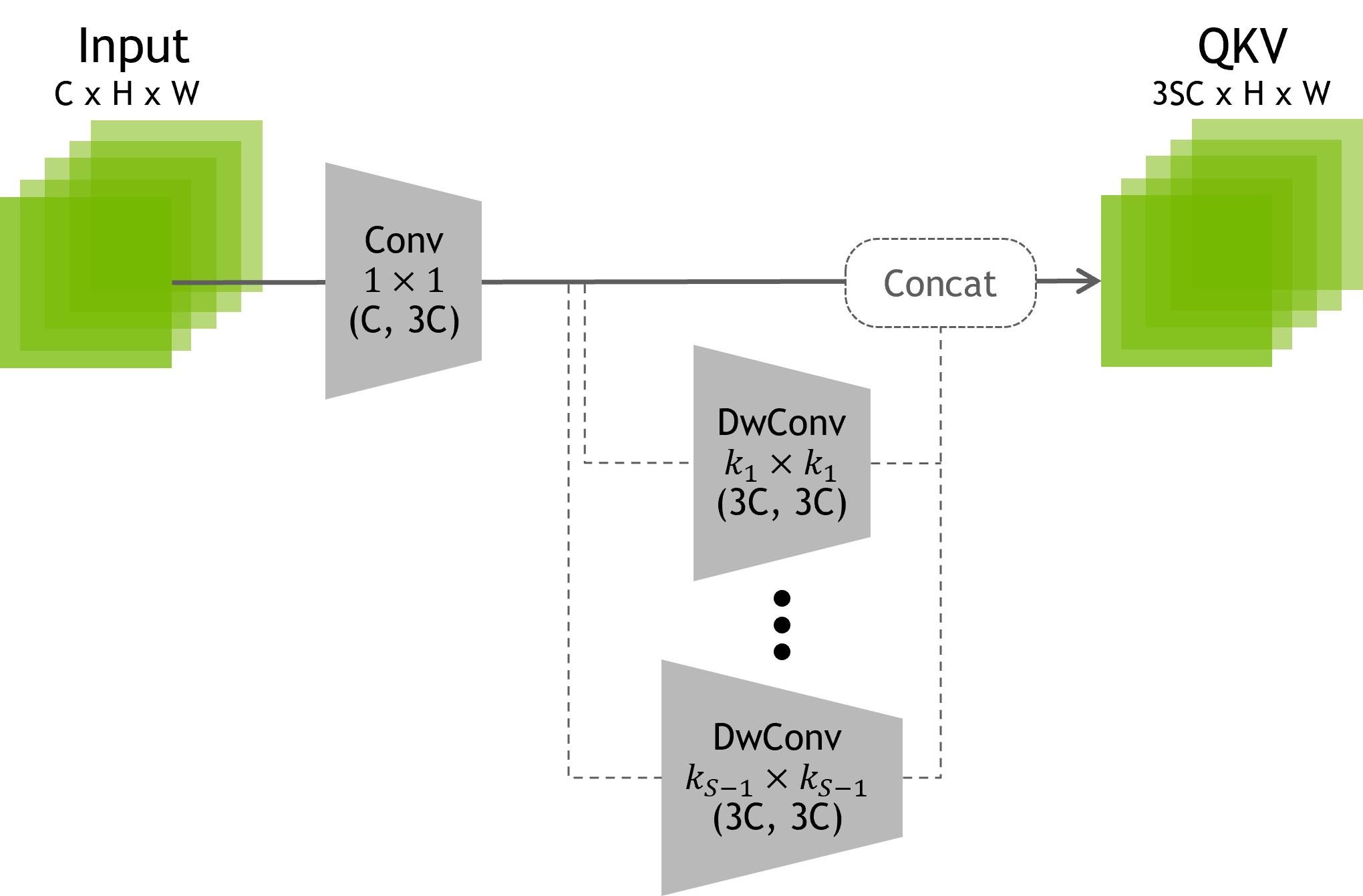}
\vspace{-1mm}
\caption{Multi-Scale Local Feature Encoding. Before upcoming block that performs attention in a global manner, this phase allows learning local contexts. }
\vspace{-5mm}
\centering
\label{fig:MSLCE}
\end{figure}

\subsection{Multi-Scale Local Context}
\john{Addressing the limitation of ReLU-based attention, which produces concentrated attention maps and struggles with local information capture \cite{cai2023efficientvit, katharopoulos2020transformers}, our method employs depth-wise convolution operators to extract local context without significantly increasing the model's parameter footprint. While depth-wise convolutions are preferred for their efficiency, alternative convolution forms are also applicable. For clarity of notations, \(C\), \(H\), and \(W\) denote the tensor dimensions of channel, height, and width, respectively. 
As depicted in Figure~\ref{fig:MSLCE}, the initial 1$\times$1 convolution brings up the number of the feature channels to $3C$, and the output from depth-wise convolutions is amalgamated with the original feature map into one tensor with channels of $3SC$. 
Here, \(S\) represents \(S-1\) depth-wise convolutions plus the original features, which are then concatenated concatenated. Each of the depth-wise convolution employs variable kernel sizes, allowing for adjustable control over the extent of local context captured. 
For example, employing depth-wise convolutions with 3$\times$3 and 5$\times$5 kernels alongside the original feature maps results in an output tensor dimension of \(3 \times 3 \times C \times H \times W\).  
For our experiments, we use one depth-wise convolution with 5$\times$5 kernel sizes concatenating with the original feature maps to construct \textit{QKV} tensor.
}

\begin{figure}[!t]
\centering
\includegraphics[width=\linewidth]{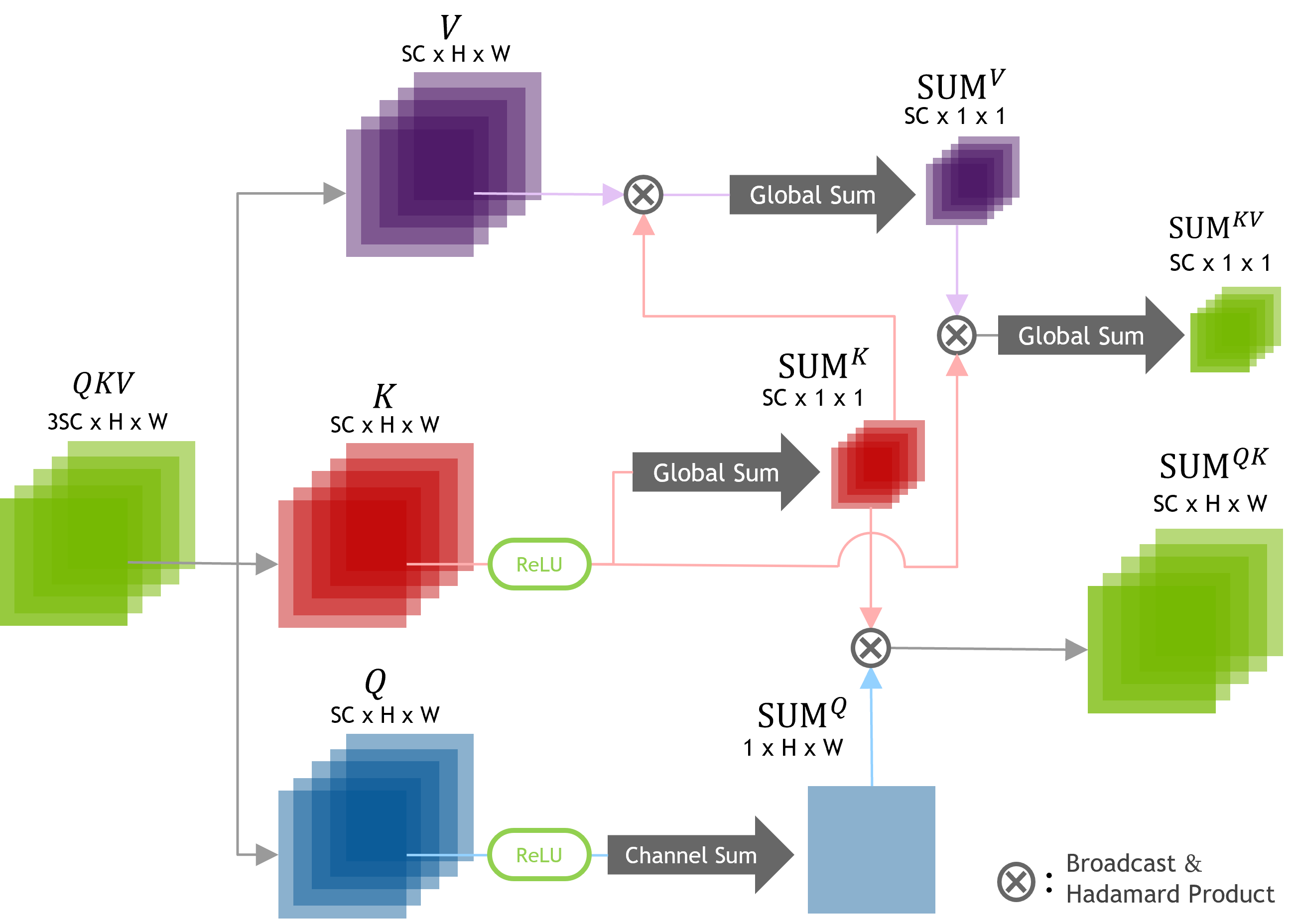}
\caption{Details of ReduceFormer Attention. The output tensors are used in Eq. \ref{eq:RF_eqn} to compute attention. } 
\vspace{-4mm}
\centering
\label{fig:RF_attn}
\end{figure}

\john{
\subsection{ReduceFormer Attention}
The idea of projecting spatial features across locations predates Transformers \cite{chang2018broadcasting, woo2018cbam}, yet the nuanced, intertwined projection achieved through self-attention modules also holds significance \cite{dosovitskiy2020image_vit}.
\textit{EfficientViT} \cite{cai2023efficientvit} performs ReLU linear attention in favor of latency over Softmax attention:
\begin{equation}
  O_i=\frac{ReLU(Q_i) \sum_{j=1}^N ReLU(K_j)^{T}V_j}{ReLU(Q_i)\sum_{j=1}^N ReLU(K_j)^{T}},
\label{eq:EV_eqn}
\vspace{-2mm}
\end{equation}
for $i$-th feature in the output feature map $O$.
The attention module allows latency improvement to some extent, but still suffers from costly computational complexity caused by matrix multiplications. 
}
\john{
To this end, 
our method utilizes repeated global summations and element-wise multiplications to globally project spatial features, aiming to approximate and bypass the inner product calculation of \textit{K} and \textit{V} from Eq. \ref{eq:EV_eqn}.
This circumvents the latency-inducing matrix multiplication inherent in conventional self-attention mechanisms, improving computational efficiency while emphasizing global feature projection. 
Furthermore, for the activation purpose, each pixel gets divided by the sum of all features.
Concretely, the computation of ReduceFormer attention block can be succinctly summarized into the following equation: 
\begin{equation}
  O_i=\frac{ReLU(Q_i )\mathbf{SUM}^{KV}}{\mathbf{SUM}_i^{QK}},  
\label{eq:RF_eqn}
\vspace{-2mm}
\end{equation}
where each term in the right-hand of Eq. \ref{eq:RF_eqn} is computed as shown in Figure \ref{fig:RF_attn}.
}
The input tensor in this module is first split into three groups of tensors $Q$, $K$ and $V$. 
$ReLU(Q_i)$ represents the $i$-th spatial feature vector from the partial input tensor $Q$ activated with \textit{ReLU}. 
Eq. \ref{eq:RF_eqn} induces the network to learn attentive features through division by the summation of all elements $\mathbf{SUM}_i^{QK}$. 
Inspired by \cite{cai2023efficientvit}, \textit{ReLU} activation is applied to $K$ and $Q$ beforehand, in order to 1) ensure their elements to be positive values when computing the division by $\mathbf{SUM}_i^{QK}$ and 2) utilize non-linear rectification while achieving linear computational complexity \cite{katharopoulos2020transformers}.

In detail, $ReLU(K)$ gets \textit{reduced} as a vector $\mathbf{SUM}^{K}\in\mathbb{R}^{SC\times1\times1}$ for  which we chose \textit{global-sum} as a succinct representation of $K$, 
allowing it to be element-wise multiplied to all spatial features of $V$.
Then, $\mathbf{SUM}^K \otimes V$ is also reduced into $\mathbf{SUM}^V \in \mathbb{R}^{SC\times1\times1}$ through a global-sum, which is followed by a element-wise multiplication with $ReLU(K)$.
This intertwined projection between spatial features allows for each element of $ReLU(K)$ and $V$ to be globally mapped onto each other, aiming to circumvent the conventional matrix multiplication between tensor elements.
Finally, $\mathbf{SUM}^{KV} \in \mathbb{R}^{SC\times1\times1}$ is created by performing another global-sum on $ReLU(K) \otimes \mathbf{SUM}^V$.
As noted in Eq~\ref{eq:RF_eqn}, this reduced representation $\mathbf{SUM}^{KV}$ is to map relational features between $ReLU(K)$ and $V$ to $ReLU(Q)$ and learn attentive relations among spatial features, followed by dividing each element with $\mathbf{SUM}_i^{QK}$ whose computation is shown in Figure~\ref{fig:RF_attn}. 

ReduceFormer attention normalizes each spatial feature in input maps, producing an output feature tensor  $O\in\mathbb{R}^{SC \times H \times W}$, where $H$ and $W$ denote the height and width, respectively. It then adjusts the features in each output $O_i$ according to weights derived from global feature relations, enhancing context learning.

\begin{table*}[!t]
\footnotesize
    \centering
	\begin{tabular}{l|cc|c|cc|c}
	\specialrule{1pt}{1pt}{1pt}
        \multirow{3}{*}{Models} & \multirow{3}{*}{\#Params} & \multirow{3}{*}{MACs} &  \multirow{2}{*}{Top1 Acc $\uparrow$}  & \multicolumn{2}{c|}{NVIDIA DRIVE Orin SoC} 
        & L40 GPU  \\ 
        &  &  & \multirow{2}{*}{(\%)} & FP16 Latency $\downarrow$ & Avg Mem BW $\downarrow$ & Throughput $\uparrow$ \\ 
        &  &  &   & (ms) & (MB/image) & (images/sec) \\
        
        \hline
        
        EfficientViT-B1 (r224)~\cite{cai2023efficientvit} & 9.1M & 0.53G & 79.4 & 0.90 & 27.48 &  3067   \\
        EfficientViT-B1 (r256)~\cite{cai2023efficientvit} & 9.1M & 0.69G & 79.9 & 0.98 & 32.96 &  2976   \\
        EfficientViT-B1 (r288)~\cite{cai2023efficientvit} & 9.1M & 0.87G & 80.4 & 1.14 & 36.84 &  2817  \\ 	 
        \textbf{ReduceFormer-B1 (r224)}                   & 9.0M & 0.52G & 79.3 & \textbf{0.68 (32\%$\downarrow$)} & \textbf{26.02 (6\%$\downarrow$)} &  \textbf{4149 (35\%$\uparrow$)} \\ 
        \textbf{ReduceFormer-B1 (r256)}                   & 9.0M & 0.67G & 80.1 & \textbf{0.73 (34\%$\downarrow$)}& \textbf{30.50 (8\%$\downarrow$)} &  \textbf{4049 (36\%$\uparrow$)} \\ 
        \textbf{ReduceFormer-B1 (r288)}                   & 9.0M & 0.85G & 80.6 & \textbf{0.87 (31\%$\downarrow$)}& \textbf{36.30 (2\%$\downarrow$)} &  \textbf{3731 (32\%$\uparrow$)} \\
        
        \hline
        
        CoAtNet-0~\cite{dai2021coatnet}                   & 25M & 4.2G & 81.6 & 2.71 & 155.25 & 1742 \\
        ConvNeXt-T~\cite{liu2022convnext}                 & 29M & 4.5G & 82.1 & 2.11 & 98.55 & 2247 \\
	EfficientViT-B2 (r256)~\cite{cai2023efficientvit} & 24M & 2.1G & 82.7 & 1.86 & 
        84.38 & 1931  \\ 
        EfficientViT-B2 (r288)~\cite{cai2023efficientvit} & 24M & 2.7G & 83.1 & 2.23 & 106.87 & 1815  \\ 
        \textbf{ReduceFormer-B2 (r256)}                   & 24M & 2.1G & 82.6 & \textbf{1.41 (32\%$\downarrow$)}&  \textbf{79.14 (7\%$\downarrow$)}& \textbf{2625 (36\%$\uparrow$)}  \\
        \textbf{ReduceFormer-B2 (r288)}                   & 24M & 2.8G & 83.0 & \textbf{1.68 (33\%$\downarrow$)}& \textbf{98.75 (8\%$\downarrow$)}& \textbf{2439 (34\%$\uparrow$)}  \\
        
        \hline
        
        Swin-B~\cite{liu2021swin}                         & 88M & 15G  & 83.5 & 4.20 & 319.98 & 1142  \\
        CoAtNet-1~\cite{dai2021coatnet}                   & 42M & 8.4G & 83.3 & 4.65 & 258.47 & 980 \\
        ConvNeXt-S~\cite{liu2022convnext}                 & 50M & 8.7G & 83.1 & 4.34 & 209.88 & 1274  \\
        EfficientViT-B3 (r224)~\cite{cai2023efficientvit} & 49M & 4.0G & 83.5 & 2.93 & 152.77  & 1267  \\ 
        EfficientViT-B3 (r256)~\cite{cai2023efficientvit} & 49M & 5.2G & 83.8 & 3.26 & 186.78 & 1203  \\ 
        \textbf{ReduceFormer-B3 (r224)}                   & 48M & 3.9G & 83.4 & \textbf{2.22 (31\%$\downarrow$)}& \textbf{138.65 (10\%$\downarrow$)} &  \textbf{1742 (37\%$\uparrow$)}  \\ 
        \textbf{ReduceFormer-B3 (r256)}                   & 48M & 5.1G & 83.6 & \textbf{2.43 (33\%$\downarrow$)}& \textbf{173.36 (8\%$\downarrow$)} &  \textbf{1631 (36\%$\uparrow$)}  \\ 

        \hline
        CoAtNet-2~\cite{dai2021coatnet}                   & 75M & 16G  & 84.1 & 6.02 & 434.55  & 845\\
        ConvNeXt-B~\cite{liu2022convnext}                 & 89M & 15G  & 83.8 & 5.82 & 351.39  & 1021 \\
        EfficientViT-B3 (r288)~\cite{cai2023efficientvit} & 49M & 6.6G & 84.2 & 4.10  & 226.43   & 1087 \\ 
        \textbf{ReduceFormer-B3 (r288)}                   & 48M & 6.4G & 84.2 & \textbf{3.03 (37\%$\downarrow$)} & \textbf{210.89 (7\%$\downarrow$)}&  \textbf{1464 (35\%$\uparrow$)}  \\ 
        
	\hline
        
        \hline
    \end{tabular}
    \vspace{-1mm}
    \caption{Classification Results on ImageNet-1K \cite{deng2009imagenet} data. Latency and throughput were measured with TensorRT in FP16 precision on both NVIDIA DRIVE Orin and L40 GPU. Memory bandwidth (Mem BW) is derived from memory read and written during inference per image and averaged over 1000 runs. The percentage in parentheses is calculated with respect to its counterpart from EfficientViT~\cite{cai2023efficientvit}.}
    \vspace{-2mm}
   \label{tab:imagenet}
\end{table*}

\begin{table}[!h]
    \footnotesize
    \centering\resizebox{\linewidth}{!}{
	\begin{tabular}{l|ccc}
	\specialrule{1pt}{1pt}{1pt}
 	\multirow{2}{*}{Models}          & \multicolumn{3}{c}{Throughput  (images/s) $\uparrow$}   \\ 
        & bs8 & bs16 & bs32    \\ \hline
        
	      E.ViT-B1 (r224)~\cite{cai2023efficientvit}  & 14084 & 19607 & 23357   \\
        E.ViT-B1 (r256)~\cite{cai2023efficientvit}  & 12841 & 16949 & 19184  \\
        E.ViT-B1 (r288)~\cite{cai2023efficientvit}  & 10974 & 13640 & 14420  \\
        \textbf{RF-B1 (r224) } & \textbf{20202 (43\%$\uparrow$)} & \textbf{28120 (43\%$\uparrow$)} &\textbf{ 32128 (38\%$\uparrow$) } \\
        \textbf{RF-B1 (r256) } &\textbf{18433 (44\%$\uparrow$)} & \textbf{24279 (43\%$\uparrow$)} & \textbf{26251 (37\%$\uparrow$)}  \\
        \textbf{RF-B1 (r288) } & \textbf{15504 (41\%$\uparrow$)}&\textbf{19093 (40\%$\uparrow$)} & \textbf{19070 (32\%$\uparrow$)}  \\
        \hline
        E.ViT-B2 (r224)~\cite{cai2023efficientvit}  & 7881 & 9864 & 10873  \\
        E.ViT-B2 (r256)~\cite{cai2023efficientvit}  & 6866 & 8285 & 7860  \\
        E.ViT-B2 (r288)~\cite{cai2023efficientvit}  & 5738 & 6554 & 5851  \\
        \textbf{RF-B2 (r224)}  & \textbf{11189 (42\%$\uparrow$)} & \textbf{13389 (36\%$\uparrow$)} & \textbf{13937 (28\%$\uparrow$)} \\
        \textbf{RF-B2 (r256)}  & \textbf{9581 (40\%$\uparrow$)} & \textbf{11276 (36\%$\uparrow$)} & \textbf{10464 (33\%$\uparrow$)} \\        
        \textbf{RF-B2 (r288)}  & \textbf{7882 (37\%$\uparrow$)}& \textbf{8748 (34\%$\uparrow$)}& \textbf{7464 (28\%$\uparrow$)} \\
        \hline
        E.ViT-B3 (r224)~\cite{cai2023efficientvit}  & 4412 & 5313  & 5209 \\
        E.ViT-B3 (r256)~\cite{cai2023efficientvit}  & 3891 & 4366 & 3735  \\
        E.ViT-B3 (r288)~\cite{cai2023efficientvit}  & 3104 & 3319 & 2769  \\
        \textbf{RF-B3 (r224)}  & \textbf{6088 (38\%$\uparrow$)} & \textbf{7201 (36\%$\uparrow$)} & \textbf{6798 (30\%$\uparrow$)}  \\
        \textbf{RF-B3 (r256)}  & \textbf{5249 (35\%$\uparrow$)} & \textbf{5848 (34\%$\uparrow$)} & \textbf{5012 (34\%$\uparrow$)}  \\
        \textbf{RF-B3 (r288)}  & \textbf{4177 (35\%$\uparrow$)} & \textbf{4392 (32\%$\uparrow$)} & \textbf{3717 (34\%$\uparrow$)}  \\
 	\hline

        \hline
    \end{tabular}}
    \vspace{-2mm}
    \caption{Throughput comparison between EfficientViT~\cite{cai2023efficientvit} (E.ViT) and ReduceFormer (RF) with different batch sizes. Measured on L40 GPU with TensorRT in FP16 precision.}
    \vspace{-4mm}
    \label{table:batch}
\end{table}

\section{Experiments}

\noindent\textbf{Implementation Details.}
Our models were implemented in PyTorch and trained with 8 NVIDIA A100 GPUs. Adam optimizer with a base learning rate of $4\times10^{-4}$ is used for training with 20 warm-up epochs with cosine learning rate schedule. The batch size for training was 128. We propose three variants of ReduceFormer: 
\begin{itemize}
    \item \textbf{B1:} Channels of $[16, 32, 64, 128, 256]$ for Stem Conv and each of the $N=4$ stages, and the depths for each stage are defined as $\{L_1, L_2, L_3, L_4\}=\{2, 3, 3, 4\}$. 
    \item \textbf{B2:} Channels of $[24, 48, 96, 192, 384]$ for Stem Conv and each of the $N=4$ stages, and the depths for each stage are defined as $\{L_1, L_2, L_3, L_4\}=\{3, 3, 5, 7\}$. 
    \item \textbf{B3:} Channels of $[32, 64, 128, 256, 512]$ for Stem Conv and each of the $N=4$ stages, and the depths for each stage are defined as $\{L_1, L_2, L_3, L_4\}=\{4, 6, 6, 9\}$. 
\end{itemize}
For all variants, ReduceFormer blocks are applied in the last two stages. 

\noindent\textbf{Performance on ImageNet.}
We benchmark on the ImageNet-1K dataset~\cite{deng2009imagenet} and present results in Table~\ref{tab:imagenet}, focusing on comparisons with EfficientViT \cite{cai2023efficientvit} on two platforms: NVIDIA DRIVE Orin SoC\footnote{\url{https://developer.nvidia.com/drive/agx}}, an embedded platform, and the L40 GPU. For DRIVE Orin, we assess latency and average memory bandwidth, measured in milliseconds and MB/image, respectively, underscoring the importance of memory bandwidth during inference on devices with limited memory. On the L40 data center GPU, we report throughput as the primary metric. In both cases, performance was measured using TensorRT\footnote{\url{https://developer.nvidia.com/tensorrt}} in FP16 precision.

Benchmark results, shown in Table~\ref{tab:imagenet}, reveal that ReduceFormer and EfficientViT achieve similar accuracy for models of comparable sizes. ReduceFormer excels in smaller variants like B1 and matches EfficientViT in larger variants. Notably, ReduceFormer offers significant runtime advantages, outperforming EfficientViT variants by an average of 32\% in inference latency on the DRIVE Orin platform and maintaining a smaller memory footprint. These performance gains are crucial for latency-sensitive applications like autonomous driving on embedded platforms, where resource constraints are significant.

On L40 GPU, ReduceFormer significantly outperforms other methods in terms of throughput for all variants at a batch size of one. Throughput comparisons between EfficientViT and ReduceFormer variants for larger batch sizes of 8, 16, and 32 are detailed in Table~\ref{table:batch}. 
ReduceFormer variants on average had a 38\% performance increase compared to the EfficientVit networks, notably achieving 44\% higher throughput for B1 variant (r-256) model comparison at a batch size of 8.
This superior performance of ReduceFormer is especially advantageous for cloud computing scenarios that require high throughput at inference.

\section{Conclusion}

In this paper, we introduced \textit{ReduceFormer}, an efficient vision model family that only utilizes basic operations to overcome the complexities of conventional attention mechanism in transformer models. By refining global context learning, our method cuts computational and memory demands, boosting efficiency for deployment on diverse platforms, from resource-constrained edge devices to data center GPUs. Going forward, we aim to adapt our models for other vision tasks.
{
    \small
    \bibliographystyle{ieeenat_fullname}
    \bibliography{main}
}


\end{document}